%% file: main.tex
\documentclass[sigconf]{acmart}

\usepackage{url}

\usepackage{multirow}
\usepackage{balance}
\usepackage{amsfonts}
\usepackage{amsmath}
\usepackage{amsthm}
\usepackage{graphicx}
\usepackage{microtype}
\usepackage{url}
\usepackage{enumitem}
\usepackage{wrapfig,lipsum}
\usepackage{multirow}
\usepackage{makecell}
\usepackage{wrapfig}
\usepackage{tabularx}
\usepackage{subfigure}
\usepackage[noend,ruled]{algorithm2e}
\usepackage{bbm}
\usepackage{dsfont}
\usepackage{algpseudocode}
\usepackage{etoolbox}
\usepackage{subfigure}
\usepackage{float}
\usepackage{booktabs}
\usepackage{caption}
\usepackage{setspace}

\newcommand{\bs}[1]{\boldsymbol{#1}}
\DeclareMathOperator*{\argmax}{arg\,max}

\renewcommand{\textrightarrow}{$\rightarrow$}

\newcommand{\ie}{\emph{i.e.}}
\newcommand{\eg}{\emph{e.g.}}

\newcommand{\nop}[1]{}

\newcommand{\josh}{\mbox{\sf JoSH}\xspace}

\theoremstyle{definition}
\newtheorem{definition}{Definition}

\AtBeginDocument{%
  \providecommand\BibTeX{{%
    \normalfont B\kern-0.5em{\scshape i\kern-0.25em b}\kern-0.8em\TeX}}}

\copyrightyear{2020}
\acmYear{2020}
\setcopyright{acmcopyright}\acmConference[KDD '20]{Proceedings of the 26th ACM SIGKDD Conference on Knowledge Discovery and Data Mining}{August 23--27, 2020}{Virtual Event, CA, USA}
\acmBooktitle{Proceedings of the 26th ACM SIGKDD Conference on Knowledge Discovery and Data Mining (KDD '20), August 23--27, 2020, Virtual Event, CA, USA}
\acmPrice{15.00}
\acmDOI{10.1145/3394486.3403242}
\acmISBN{978-1-4503-7998-4/20/08}

\begin{document}

\fancyhead{}

\leftmargini=12pt 
\title{Hierarchical Topic Mining via Joint Spherical Tree and Text Embedding}


\author{Yu Meng$^{1*}$, Yunyi Zhang$^{1*}$, Jiaxin Huang$^{1}$, Yu Zhang$^{1}$, Chao Zhang$^{2}$, Jiawei Han$^{1}$}
\affiliation{
\institution{$^1$Department of Computer Science, University of Illinois at Urbana-Champaign, IL, USA} 
\institution{$^2$College of Computing, Georgia Institute of Technology, GA, USA}
\institution{$^{1}$\{yumeng5, yzhan238, jiaxinh3, yuz9, hanj\}@illinois.edu \ \ \ $^2$chaozhang@gatech.edu}
}
\thanks{$^*$Equal Contribution.}



\input{0-abs}



\maketitle

\input{1-intro}

\input{2-def}

\input{3-emb}

\input{4-opt}

\input{5-sum}

\input{6-exp}

\input{7-related}

\input{8-concl}

\begin{acks}
Research was sponsored in part by US DARPA KAIROS Program No. FA8750-19-2-1004 and SocialSim Program No.  W911NF-17-C-0099, National Science Foundation IIS 16-18481, IIS 17-04532, and IIS 17-41317, and DTRA HDTRA11810026. Any opinions, findings, and conclusions or recommendations expressed herein are those of the authors and should not be interpreted as necessarily representing the views, either expressed or implied, of DARPA or the U.S. Government. The U.S. Government is authorized to reproduce and distribute reprints for government purposes notwithstanding any copyright annotation hereon. 
We thank anonymous reviewers for valuable and insightful feedback.
\end{acks}


\bibliographystyle{ACM-Reference-Format}
\balance
\bibliography{ref}

\input{9-app}
\end{document}

%% file: 0-abs.tex

\begin{abstract}
Mining a set of meaningful topics organized into a hierarchy is intuitively appealing since topic correlations are ubiquitous in massive text corpora.
To account for potential hierarchical topic structures, hierarchical topic models generalize flat topic models by incorporating latent topic hierarchies into their generative modeling process.
However, due to their purely unsupervised nature, the learned topic hierarchy often deviates from users' particular needs or interests. 
To guide the hierarchical topic discovery process with minimal user supervision, we propose a new task, Hierarchical Topic Mining, which takes a category tree described by category names only, and aims to mine a set of representative terms for each category from a text corpus to help a user comprehend his/her interested topics.
We develop a novel joint tree and text embedding method along with a principled optimization procedure that allows simultaneous modeling of the category tree structure and the corpus generative process in the spherical space for effective category-representative term discovery. 
Our comprehensive experiments show that our model, named \josh, mines a high-quality set of hierarchical topics with high efficiency and benefits weakly-supervised hierarchical text classification tasks\footnote{Source code can be found at \url{https://github.com/yumeng5/JoSH}.}. 
\end{abstract}

%% file: 1-intro.tex

\section{Introduction}
Topic models~\cite{Blei2003LatentDA,Hofmann1999ProbabilisticLS}, which uncover hidden semantic structure in a text corpus via generative modeling, have proven successful on automatic topic discovery. Hierarchical topic models~\cite{Blei2003HierarchicalTM,Mimno2007MixturesOH} extend the classical ones by considering a latent topic hierarchy during the corpus generative process, motivated by the fact that topics are naturally correlated (\eg, ``sports'' is a super-topic of ``soccer''). Due to their effectiveness of discovering organized topic structures automatically without human supervision, hierarchical topic models have been applied to a wide range of applications including political text analysis~\cite{grimmer2010bayesian}, entity disambiguation~\cite{kataria2011entity} and relation extraction~\cite{Alfonseca2012PatternLF}.

Despite being able to learn latent topic hierarchies from text corpora, the applicability of hierarchical topic models to learn a user-interested topic structure is limited seriously by their unsupervised nature: Unsupervised generative models maximize the likelihood of the observed data, tending to discover the most general and prominent topics from a text collection, which may not fit a user's particular interest, or provide a superficial summarization of the corpus. Furthermore, the inference algorithms of topic models yield local optimum solutions, resulting in instability and inconsistency across different runs. This issue even worsens in the hierarchical setting where a larger number of topics and their correlations need to be modeled.

\begin{figure}[t]
\centering
\includegraphics[width=\linewidth]{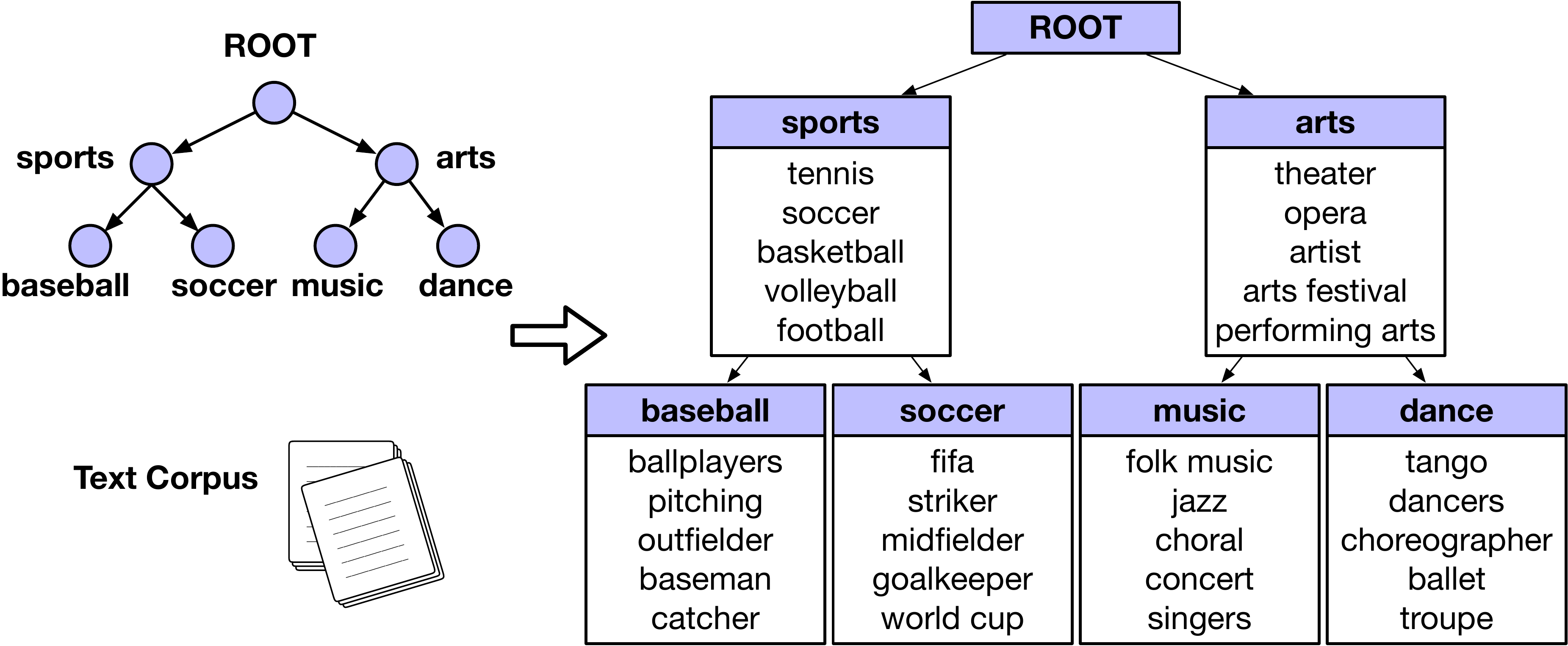}
\caption{An example of Hierarchical Topic Mining. We aim to retrieve a set of representative terms from a given corpus for each category in a user-provided hierarchy.}
\label{fig:eg}
\end{figure}
In many cases, a user is interested in a specific topic structure, or has prior knowledge about the potential topics in a corpus. 
These topics, based on a user's interest or prior knowledge, may be easily described via a set of category names with a hierarchical structure.
Such a user-provided category hierarchy will facilitate a more stable topic discovery process, yielding more desirable and consistent results that better cater to a user's needs. 
Therefore, we propose a new task, \textbf{Hierarchical Topic Mining}, which takes only a topic hierarchy described by category names as user guidance, and aims to retrieve a set of coherent and representative terms under each category to help users comprehend his/her interested topics. For example, as shown in Figure~\ref{fig:eg}, a user may provide a hierarchy of interested concepts along with a corpus and rely on hierarchical topic mining to retrieve a set of representative terms from a text corpus (\eg, different music and dance genres, terminologies for different sports, as well as general descriptions for internal nodes) that provide a clear interpretation of the categories. 

Several previous studies also focus on guiding topic discovery with word-level supervision. Seed-guided topic modeling~\cite{Andrzejewski2009LatentDA,Jagarlamudi2012IncorporatingLP} incorporates user-provided seed words to bias the generative process towards seed-related topics. A recent study CatE~\cite{Meng2020DiscriminativeTM} learns discriminative text embeddings guided by category names for representative term retrieval. However, none of the above methods handle hierarchical topic structures. Under the hierarchical setting, there are supervised~\cite{Perotte2011HierarchicallySL} and semi-supervised~\cite{Mao2012SSHLDAAS} models that leverage category labels of documents to regularize the generative process. However, they rely on a large amount of annotated documents which may be costly to obtain. Under our setting, only a set of easy-to-provide category names that form a topic hierarchy is needed to guide the hierarchical topic discovery process.

In this paper, we propose \josh, a novel \textbf{Jo}int \textbf{S}pherical tree and text embedding model for \textbf{H}ierarchical Topic Mining. The user-provided category tree structure and text corpus statistics are simultaneously modeled via directional similarity in the spherical space, which facilitates effective estimation of category-word semantic correlations for representative term discovery. To train our model in the spherical space, we develop a principled EM optimization procedure based on Riemannian optimization.

Our contributions can be summarized as follows.
\begin{enumerate}
\item We propose a new task for hierarchical topic discovery, Hierarchical Topic Mining, which requires a category hierarchy described by category names as the only supervision to retrieve a set of representative terms per category for effective topic understanding.

\item We develop a joint embedding framework for hierarchical topic mining by simultaneously modeling the user-provided category tree structure and the text generation process. The model is defined in the spherical space, where directional similarity is employed to characterize semantic correlations among words, documents, and categories for accurate category representative term retrieval. 

\item We develop an EM algorithm to optimize our model in the spherical space that iterates between estimating the latent category of words and maximizing corpus generative likelihood while optimizing the category tree structure in the embedding space.

\item We conduct a comprehensive set of experiments on two public corpora from different domains on Hierarchical Topic Mining. Our model enjoys high efficiency and mines high-quality topics. The embeddings trained by our model can be directly used for weakly-supervised hierarchical text classification.

\end{enumerate}

%% file: 2-def.tex

\section{Problem Formulation}
\begin{definition} [Hierarchical Topic Mining]
Given a text corpus $\mathcal{D}$ and a tree-structured
hierarchy $\mathcal{T}$ where each node $c_i \in \mathcal{T}$ is represented by the name of the category, \textbf{Hierarchical Topic Mining} aims to retrieve a set of terms $\mathcal{C}_i=\{w_1, \dots, w_m\}$ from $\mathcal{D}$ for each category $c_i \in \mathcal{T}$ such that $\mathcal{C}_i$ provides a clear description of the category $c_i$ based on $\mathcal{D}$.
\end{definition}

\noindent 
\textbf{Connection and difference between Hierarchical Topic Models.} Similar to Hierarchical Topic Modeling~\cite{Blei2003HierarchicalTM}, we also aim to capture the hierarchical correlations among topics during topic discovery. However, \textbf{Hierarchical Topic Mining} is weakly-supervised as it requires the user to provide the names of the hierarchy categories which serve as the minimal supervision and focuses on retrieving representative terms only for the provided categories.

%% file: 3-emb.tex

\section{Spherical Text and Tree Embedding}
\label{sec:tree_text_emb}

In this section, we introduce our model \josh which jointly learns text embeddings and tree embeddings in the spherical space, where directional similarity is used to effectively characterize semantic correlations among words, documents and categories. 

\subsection{Motivation}
Mining representative terms relevant to a given category relies on accurate estimation of semantic similarity, on which \emph{directional similarity} of text embeddings has proven most effective.
For example, cosine similarity is empirically shown~\cite{Levy2014LinguisticRI} to better characterize word semantic similarity and dissimilarity.
Motivated by the effectiveness of directional similarity for text analysis, several recent studies employ the spherical space for topic modeling~\cite{batmanghelich2016nonparametric}, text embedding learning~\cite{meng2019spherical} and text sequence generation~\cite{Kumar2019VonML}.
To learn text embeddings tailored for the given category tree, we propose to jointly embed the tree structure into the spherical space where each category is surrounded by its representative terms.


Different from recent hyperbolic tree embedding models, such as Poincar{\'e} embedding~\cite{Nickel2017PoincarEF}, Lorentz model~\cite{Nickel2018LearningCH} and hyperbolic cones~\cite{Ganea2018HyperbolicEC}, we do not preserve the \emph{absolute} tree distance in the embedding space, but rather the \emph{relative} category relationship reflected in the tree structure. For example, in the category hierarchy given in Figure~\ref{fig:eg}, although the tree distance between ``sports'' and ``arts'' and that between ``baseball'' and ``soccer'' are both $2$, the latter pair of categories should be embedded closer than the former pair due to higher semantic similarity. Therefore, the tree distance in the category hierarchy should not be preserved in an absolute manner, but treated as a relative metric, \eg, for category ``soccer'', its tree distance to ``sports'' is smaller than that to ``baseball'', so ``soccer'' should be embedded closer to ``sports'' than to ``baseball''.

\subsection{Spherical Tree Embedding}
\begin{figure*}[ht]
\subfigure[Intra- \& Inter-Category Configuration.]{
\label{fig:intra_inter}
\includegraphics[width = 0.265\textwidth]{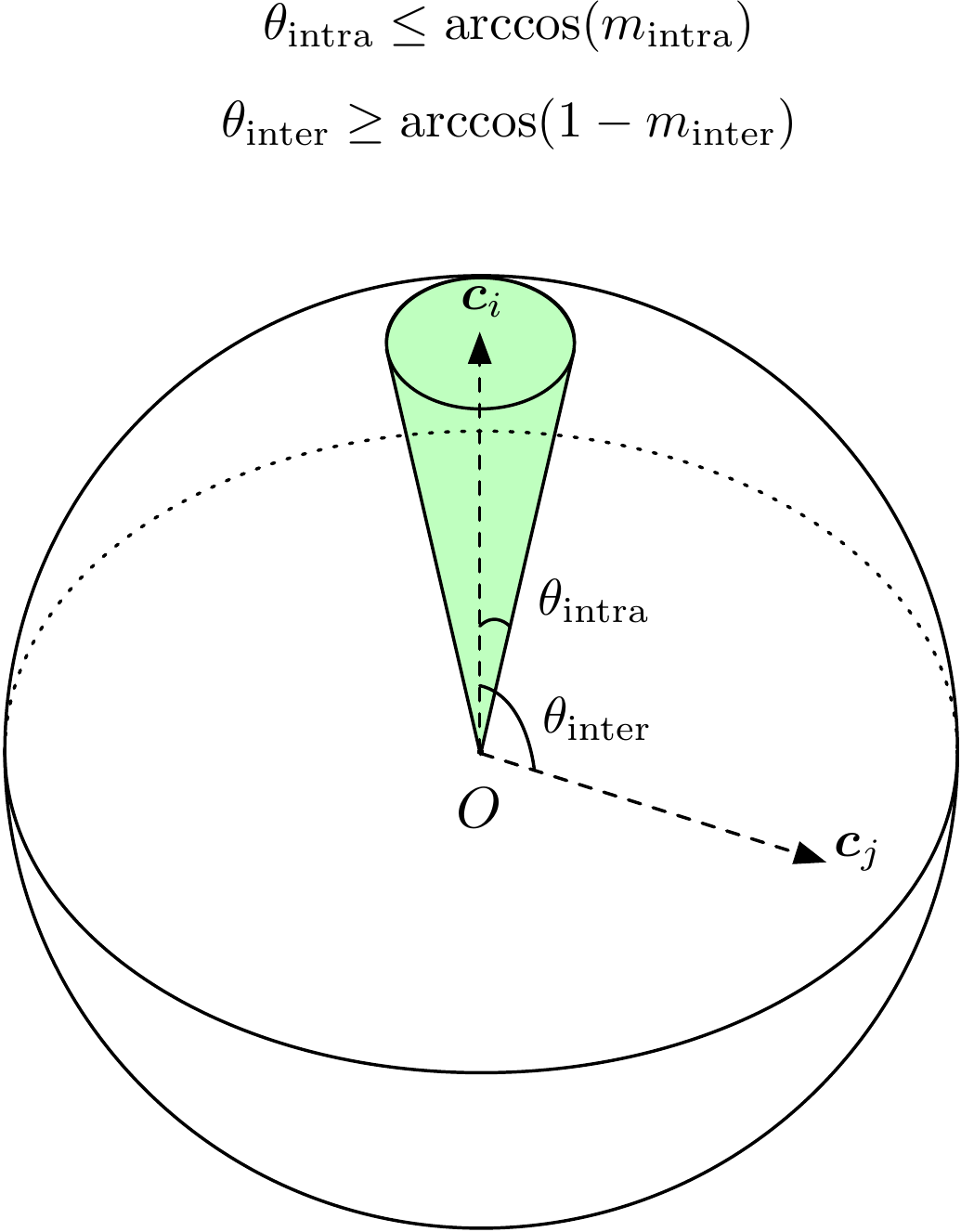}
}
\subfigure[Embed First-Level Local Tree.]{
\label{fig:emb_1}
\includegraphics[width = 0.34\textwidth]{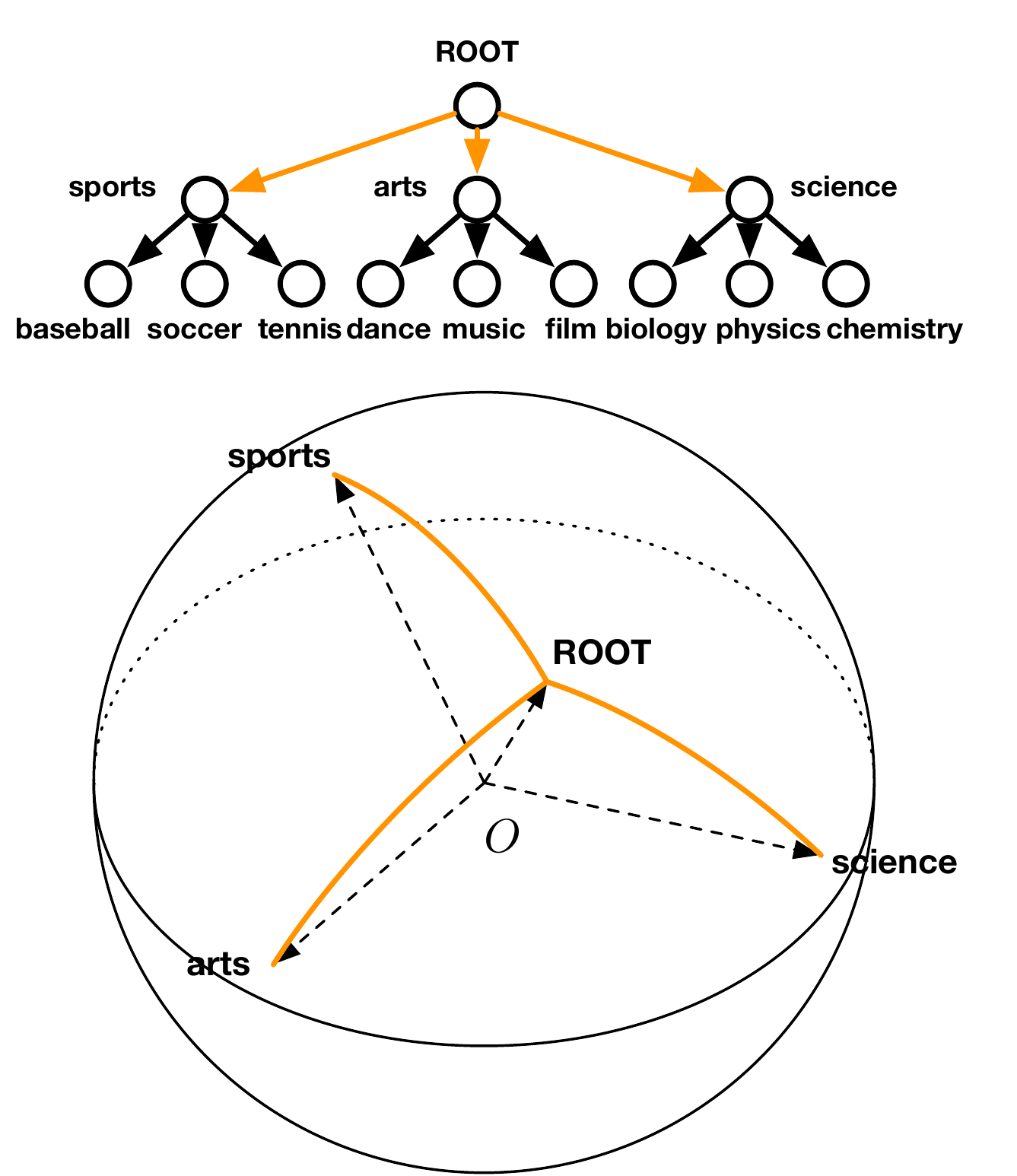}
}
\subfigure[Embed Second-Level Local Trees.]{
\label{fig:emb_2}
\includegraphics[width = 0.34\textwidth]{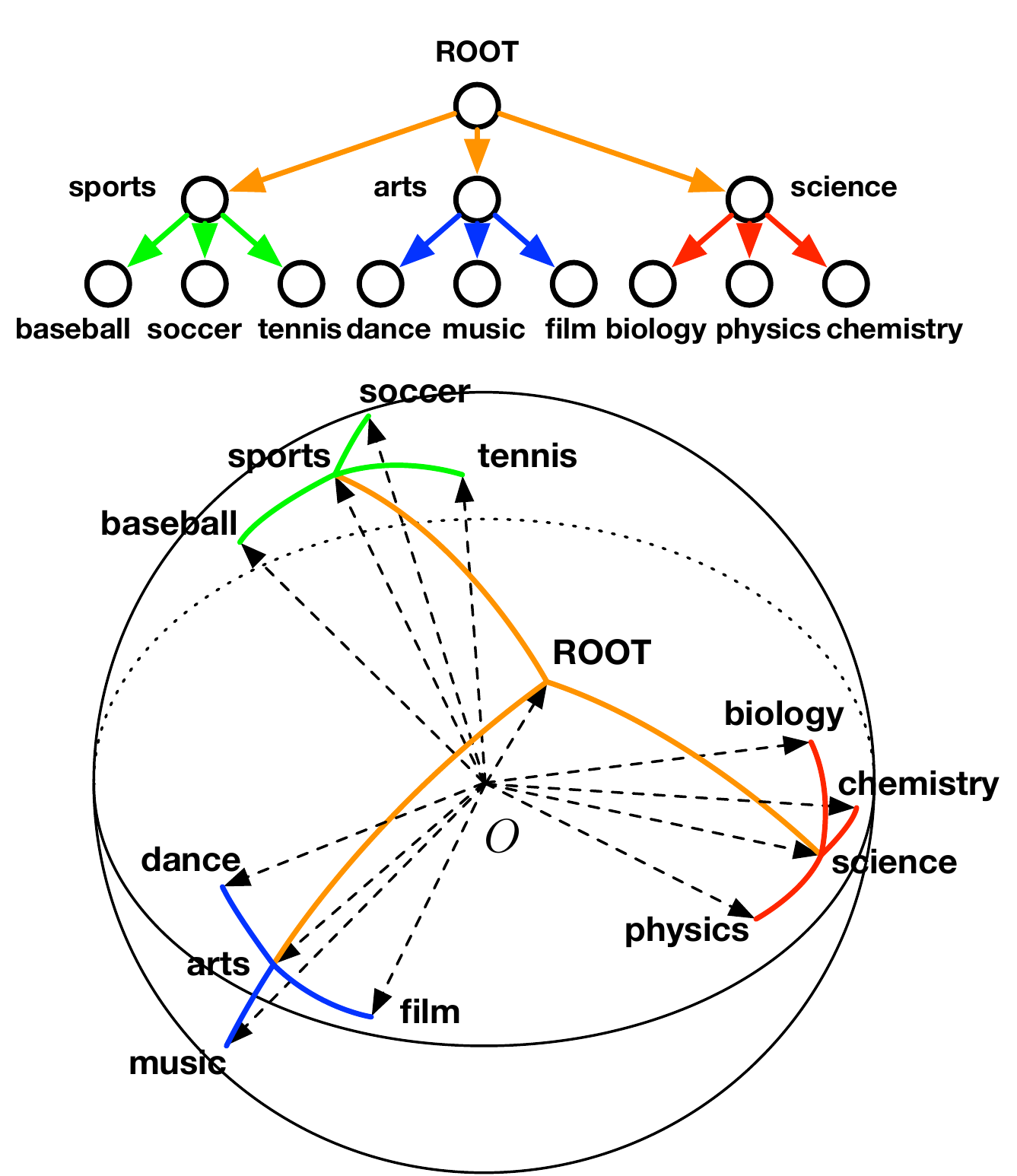}
}
\caption{Spherical tree embeddings. All category center vectors reside on the unit sphere. (a) Representative terms are pushed into a spherical sector centered around the category center vector. Directional distance is enforced between categories. (b) \& (c) Local trees are recursively embedded onto the sphere.}
\label{fig:sphere}
\end{figure*}
We propose a novel tree embedding method that preserves the relative category hierarchical structure in the spherical embedding space, meanwhile encouraging inter-category distinctiveness for clear topic interpretation.

\subsubsection{The Flat Case}

We start with the simplest case where all categories are parallel and do not exhibit hierarchical structures. We aim to jointly embed categories and their representative terms such that (1) the representative terms selected for each category\footnote{We will discuss how to select representative terms in Section~\ref{sec:opt}.} are semantically coherent and (2) the categories are distinctive from each other, which allows clear category interpretation. For example, in Figure~\ref{fig:eg}, one can clearly recognize and understand ``baseball'' and ``soccer'' thanks to the discriminative terms that are exclusively relevant to the corresponding category.

\noindent
\textbf{Intra-Category Coherence.} 
The representative terms of each category should be highly semantically relevant to each other, reflected by high directional similarity in the spherical space. To achieve this, we require the embeddings of representative terms to be placed near the category center direction within a local region by maximizing
\begin{equation}
\label{eq:obj_intra}
\mathcal{L}_{\text{intra}} = \sum_{c_i \in \mathcal{T}} \sum_{w_j \in \mathcal{C}_i} \min(0, \bs{u}_{w_j}^\top\bs{c}_i - m_{\text{intra}}),
\end{equation}
where $\bs{u}_{w}$ is the word embedding of $w$; $\bs{c}_i$ is the category center vector of $c_i$. Note that $\bs{u}_{w_j}^\top\bs{c}_i=\cos(\bs{u}_{w_j}, \bs{c}_i)$ since the vectors reside on the unit sphere $\mathbb{S}^{p-1}\subset \mathbb{R}^p$.
We set $m_{\text{intra}}=0.9$ which works well in general since it requires high cosine similarity between representative words and the category center.

When $\mathcal{L}_{\text{intra}}$ is maximized (\ie, $\forall w_j \in \mathcal{C}_i, \bs{u}_{w_j}^\top \bs{c}_i \ge m_{\text{intra}}$), the representative word embeddings of the corresponding category reside in a spherical sector centered around the category center vector.

\noindent
\textbf{Inter-Category Distinctiveness.}
We would like to encourage distinctiveness across different categories to avoid semantic overlaps so that the retrieved terms provide a clear and distinctive description of the category. To accomplish this, we enforce inter-category directional dissimilarity by requiring the cosine distance between any two categories to be larger than $m_{\text{inter}}$, \ie,
$$
\forall c_i, c_j (c_i \neq c_j), 1 - \bs{c}_i^\top \bs{c}_j > m_{\text{inter}}.
$$
Therefore, we maximize the following objective:
\begin{equation}
\label{eq:obj_inter}
\mathcal{L}_{\text{inter}} = \sum_{c_i\in \mathcal{T}}  \sum_{c_j \in \mathcal{T}\setminus \{c_i\}} \min(0, 1 - \bs{c}_i^\top \bs{c}_j - m_{\text{inter}}).
\end{equation}
We will introduce how to set $m_{\text{inter}}$ in Section~\ref{sec:recursive}.

Figure~\ref{fig:intra_inter} shows the configuration of category center vectors upon enforcing intra-category coherence and inter-category distinctiveness.

\subsubsection{Recursive Local Tree Embedding}
\label{sec:recursive}
We generalize the ideas in the flat case to the hierarchical case and recursively embed local structures of the category tree such that the relative category relationship is preserved.

We first define the local tree structure that we work with at each recursive step:
\begin{definition} [Local Tree]
A local tree $\mathcal{T}_r$ rooted at node $c_r \in \mathcal{T}$ consists of node $c_r$ and all its \emph{direct} children nodes.
\end{definition}

\noindent
\textbf{Preserving Relative Tree Distance Within Local Trees.}
Without a hierarchical structure, pairwise category distance is enforced by Eq.~\eqref{eq:obj_inter}. With a local tree structure, the category distance in the embedding space should reflect the tree distance in a \emph{comparative} way. Specifically, since the tree distance between two children nodes is larger than that between a children node and the parent node, a category should be closer to its parent category than to its sibling categories in the embedding space. To achieve this property, we employ the following objective for categories in a local tree $\mathcal{T}_r$:
\begin{equation}
\label{eq:obj_dis_mod}
\mathcal{L}_{\text{inter}} = \sum_{c_i \in \mathcal{T}_r\setminus \{c_r\}} \sum_{c_j \in \mathcal{T}_r\setminus \{c_r, c_i\}} \min(0, \bs{c}_i^\top \bs{c}_r - \bs{c}_i^\top \bs{c}_j - m_{\text{inter}}),
\end{equation}
which generalizes Eq.~\eqref{eq:obj_inter} by forcing the directional similarity between a children category center vector and its parent category center vector to be higher than that between two sibling categories by $m_{\text{inter}}$.

Maximizing $\mathcal{L}_{\text{inter}}$ results in two favorable tree embedding properties: (1) The children categories are placed near the parent category (by requiring higher value of $\bs{c}_i^\top \bs{c}_r$), which reflects the semantic correlation between a sub-category and a super-category; (2) Any two sibling categories are well-separated (by requiring lower value of $\bs{c}_i^\top \bs{c}_j$), which encourages distinction between sibling categories (\eg, ``baseball'' vs. ``soccer'').

\noindent
\textbf{Recursively Embed Local Trees.}
We apply the idea of local tree embedding recursively to embed the entire category tree structure in a top-down manner: We first embed the local tree rooted at the ROOT node, and then proceed to the next level to embed the local trees of every node at the current level. We repeat this process until we reach the leaf nodes. Figures~\ref{fig:emb_1} and ~\ref{fig:emb_2} illustrate the recursive embedding procedure, which can be realized via the following holistic objective which combines the objectives of every local tree:
\begin{equation*}
\label{eq:obj_tree}
\mathcal{L}_{\text{tree}} = \sum_{c_r \in \mathcal{T}} \sum_{c_i \in \mathcal{T}_r\setminus \{c_r\}} \sum_{c_j \in \mathcal{T}_r\setminus \{c_r, c_i\}} \min(0, \bs{c}_i^\top \bs{c}_r - \bs{c}_i^\top \bs{c}_j - m_{\text{inter}}).
\end{equation*}

We note that $m_{\text{inter}}$ needs to be set differently for different levels: As Figure~\ref{fig:emb_2} shows, the sibling categories are embedded in more localized regions as we proceed to the lower levels of the hierarchy to reflect their intrinsic semantic similarity. As a result, for each level $L$ of $\mathcal{T}$, we set $m_{\text{inter}}(L)$ to be the average difference between children-parent and inter-sibling embedding similarity across level $L$, \ie,
$$
m_{\text{inter}}(L) = \frac{1}{N_{L}} \sum_{c_r \in L} \sum_{c_i \in \mathcal{T}_r\setminus \{c_r\}} \sum_{c_j \in \mathcal{T}_r\setminus \{c_r, c_i\}} \bs{c}_i^\top\bs{c}_r - \bs{c}_i^\top\bs{c}_j,
$$
where $N_{L}$ is the total number of sibling pairs within each local tree in level $L$. For the simplicity of notations, we omit the argument of $m_{\text{inter}}$ in the rest of the paper, but it should be kept in mind that $m_{\text{inter}}$ is level-dependent.

Finally, after embedding the category tree, we use the same objective as Eq.~\eqref{eq:obj_intra} to encourage intra-category coherence of retrieved terms so that the category embedding configuration can effectively guide the text embeddings to fit the tree structure.

\subsection{Spherical Text Embedding via Modeling Conditional Corpus Generation}
\label{sec:text_emb}
We introduce how to learn text embeddings tailored for the given  category hierarchy $\mathcal{T}$ in the spherical space by modeling the corpus generation process conditioned on the categories. Specifically, we assume the corpus $\mathcal{D}$ is generated following a three-step process: 

\begin{enumerate}[wide, labelwidth=!, labelindent=0pt]
\item First, each document $d_i \in \mathcal{D}$ is generated conditioned on one of the categories in the category hierarchy $\mathcal{T}$. 
Since a category can cover a wide range of semantics, it is natural to model a category as a distribution in the embedding space instead of as a single vector. Therefore, we extend the previous representation of a category $c_i$ from a single center vector $\bs{c}_i$ to a spherical distribution centered around $\bs{c}_i$, \ie, a von Mises-Fisher (vMF) distribution. 
Specifically, the vMF distribution of a category is parameterized by a mean vector $\bs{c}_i$ and a concentration parameter $\kappa_{c_i}$. The probability density closer to $\bs{c}_i$ is greater and the spread is controlled by $\kappa_{c_i}$. Formally, a unit random vector $\bs{x} \in \mathbb{S}^{p-1} \subset \mathbb{R}^{p}$ has the $p$-variate vMF distribution $\text{vMF}_p(\bs{x}; \bs{c}_i, \kappa_{c_i})$ if its probability density function is
\begin{equation*}
\label{eq:vmf}
f(\bs{x};\bs{c}_i, \kappa_{c_i}) = n_p(\kappa_{c_i})\exp \left( \kappa_{c_i} \cdot \cos(\bs{x}, \bs{c}_i) \right),
\end{equation*}
where $\|\bs{c}_i\| = 1$ is the center direction, $\kappa_{c_i} \ge 0$ is the concentration parameter, and the normalization constant $n_p(\kappa_{c_i})$ is given by
$$
n_p(\kappa_{c_i}) = \frac{\kappa_{c_i}^{p/2-1}}{(2\pi)^{p/2} I_{p/2-1}(\kappa_{c_i})},
$$
where $I_r(\cdot)$ represents the modified Bessel function of the first kind at order $r$. 

We define the generative probability of each document $d_i$ conditioned on its corresponding true category $c_i$ to be:
\begin{align}
\label{eq:gen_topic}
p(d_i \mid c_i) = \text{vMF}(\bs{d}_i;\bs{c}_i,\kappa_{c_i}) = n_p(\kappa_{c_i})\exp \left(\kappa_{c_i} \cdot \cos(\bs{d}_i, \bs{\bs{c}_i}) \right),
\end{align}
where $\bs{d}_i$ is the document embedding of $d_i$.

However, modeling category distribution via Eq.~\eqref{eq:gen_topic} is not directly helpful for our task, since our goal is to discover representative terms rather than documents for each category. For this reason, we further decompose $p(d_i \mid c_i)$ into category-word distribution:
\begin{equation}
\label{eq:gen_topic_mod}
p(d_i \mid c_i) \propto \prod_{w_j \in d_i} p(w_j\mid c_i) \propto \prod_{w_j \in d_i} \text{vMF}(\bs{u}_{w_j};\bs{c}_i,\kappa_{c_i}),
\end{equation}
where each word is assumed to be generated independently based on the document category.
Eq.~\eqref{eq:gen_topic_mod} allows direct modeling of $p(w_j\mid c_i)$, from which category representative terms will be derived.

\item 
Second, each word $w_j$ is generated based on the semantics of the document $d_i$. Intuitively, higher directional similarity implies higher semantic coherence, thus higher probability of co-occurrence. We assume the probability of $w_j$ appearing in document $d_i$ to be:
\begin{align}
\label{eq:gen_global}
p(w_j \mid d_i) \propto \exp(\cos(\bs{u}_{w_j}, \bs{d}_i)).
\end{align}

\item
Third, surrounding words $w_{j+k}$ in the local context window ($-h \le k \le h, k \neq 0$, $h$ is the local context window size) of $w_j$ are generated conditioned on the semantics of the center word $w_j$. Similar to (2), we assume the probability of $w_{j+k}$ appearing in the local context window of $w_j$ to be:
\begin{align}
\label{eq:gen_local}
p(w_{j+k} \mid w_j) \propto \exp(\cos(\bs{v}_{w_{j+k}}, \bs{u}_{w_j})),
\end{align}
where $\bs{v}_{w}$ is the context word representation of $w$.
\end{enumerate}

We summarize how the above three steps jointly model the text generation process by capturing both global and local textual contexts, conditioned on the given categories: Step (1) draws a connection between each document and one of the categories in $\mathcal{T}$ (\ie, \emph{topic assignment}). Step (2) models the semantic coherence between a word and the document it appears in (\ie, \emph{global contexts}). Step (3) models the semantic correlations of co-occurring words within a local context window (\ie, \emph{local contexts}). 
We note that all three steps use directional similarity to model the correlations among categories, documents, and words.

%% file: 4-opt.tex

\section{Optimization}
\label{sec:opt}

In this section, we introduce the optimization procedure for learning embedding in the spherical space via our model defined in the previous section.
\subsection{Overview}
We first summarize the objectives of our optimization problem as follows (the derivation is based on maximum likelihood estimation; details can be found at Appendix~\ref{sec:app_obj}):
$$
\mathcal{L} = \mathcal{L}_{\text{tree}} + \mathcal{L}_{\text{text}},
$$
\begin{equation}
\label{eq:obj_tree_cp}
\mathcal{L}_{\text{tree}} = \sum_{c_r \in \mathcal{T}} \sum_{c_i \in \mathcal{T}_r\setminus \{c_r\}} \sum_{c_j \in \mathcal{T}_r\setminus \{c_r, c_i\}} \min(0, \bs{c}_i^\top \bs{c}_r - \bs{c}_i^\top \bs{c}_j - m_{\text{inter}}).
\end{equation}

\begin{equation}
\begin{aligned}
\label{eq:obj_text}
\mathcal{L}_{\text{text}} &= \sum_{d_i\in \mathcal{D}} \sum_{w_j \in d_i} \sum_{\substack{w_{j+k} \in d_i \\ -h \le k \le h, k \neq 0}} \min\bigg(0,  \bs{v}_{w_{j+k}}^\top \bs{u}_{w_j} + \bs{u}_{w_j}^\top \bs{d}_i \\
& \quad  - \bs{v}_{w_{j+k}}^\top \bs{u}_{w_j'} - \bs{u}_{w_j'}^\top \bs{d}_i -m \bigg) \\
& + \sum_{c_i \in \mathcal{T}} \sum_{w_j \in \mathcal{C}_i} \left( \log \left(n_p(\kappa_{c_i})\right) + \kappa_{c_i} \bs{u}_{w_j}^\top \bs{c}_i \right) \mathds{1}(\bs{u}_{w_j}^\top \bs{c}_i < m_{\text{intra}}).
\end{aligned}
\end{equation}
$$
s.t. \quad \forall w,d,c, \quad \|\bs{u}_w\|=\|\bs{v}_w\|=\|\bs{d}\|=\|\bs{c}\|=1, \kappa_{c} \ge 0,
$$
where $\mathds{1}(\cdot)$ is the indicator function; we set $m = 0.25$.

We note that our objective contains latent variables, \ie, the second term in Eq.~\eqref{eq:obj_text} requires knowledge about the latent category of words. At the beginning, we only know that the category name provided by the user belongs to the corresponding category (\eg, $w_{\text{sports}}\in \mathcal{C}_{\text{sports}}$). The goal of Hierarchical Topic Mining is to discover the latent category assignment of more words such that they form a clear description of the category.

To solve the optimization problem involving latent variables, we develop an EM algorithm that iterates between the estimation of the latent category assignment of words (\ie, \textbf{E-Step}) and maximization of the embedding training objectives (\ie, \textbf{M-Step}). We detail the design of the EM algorithm below:

\noindent
\textbf{E-Step.}
We update the estimation of words assigned to each category by
\begin{equation}
\begin{aligned}
\label{eq:e_step}
\mathcal{C}_i^{(t)} \gets \text{Top}_t \left(\{w\}; \bs{u}_{w}^{(t)},\bs{c}_i^{(t)},\kappa_{c_i}^{(t)} \right),
\end{aligned}
\end{equation}
where $\text{Top}_t(\{w\}; \bs{u}_{w},\bs{c}_i,\kappa_{c_i})$ denotes the set of terms ranked at the top $t$ positions according to $\text{vMF}(\bs{u}_{w};\bs{c}_i,\kappa_{c_i})$ (\ie, we assign the $t$ terms to $c_i$ that are most likely generated from its current estimated category distribution). In practice, we find that gradually increasing $t$ (\ie, set $t=1$ at the first iteration where $\mathcal{C}_i^{(1)}$ contains only the category name $w_{c_i}$ by initializing $\bs{c}_i^{(1)}=\bs{u}_{w_{c_i}}^{(1)}$; increment $t$ by $1$ for the following iterations) works well. Therefore, here $t$ also denotes the iteration index.

Note here that we only update the estimation of category assignment for the top $t$ words per category, which will become the representative terms retrieved. The reason is that most of the terms in the vocabulary are not representative for any of the categories; assigning them to one of the category will have negative impact on accurate estimation of the category distribution. 

\noindent
\textbf{M-Step.}
We update the text embeddings and category embeddings by maximizing $\mathcal{L}_{\text{tree}}$ and $\mathcal{L}_{\text{text}}$:


%
%
%

\begin{equation}
\label{eq:emb}
\bs{\Theta}^{(t+1)} \gets \arg\,\max \left( \mathcal{L}_{\text{text}}\left(\bs{\Theta}^{(t)}\right) + \mathcal{L}_{\text{tree}}\left(\bs{\Theta}^{(t)}\right) \right),
\end{equation}
where $\bs{\Theta}^{(t)} = \left\{ \bs{u}_{w}^{(t)}, \bs{v}_{w}^{(t)}, \bs{d}^{(t)}, \bs{c}^{(t)} \right\}$.

Eq.~\eqref{eq:emb} requires non-Euclidean stochastic optimization methods, which will be introduced in the next subsection.

\subsection{Riemannian Optimization}
Embedding learning is usually based on stochastic optimization techniques, but Euclidean optimization methods like SGD cannot be directly applied to our case, because the Euclidean gradient provides update directions in a non-curvature space, while the embeddings in our model must be updated on the spherical surface $\mathbb{S}^{p-1}$ with constant positive curvature. 

For the above reason, we apply the Riemannian optimization method in the spherical space as described in ~\cite{meng2019spherical} to train text and tree embeddings. Specifically, the Riemannian gradient of a parameter $\bs{\theta}$ is computed as
$$
\text{grad}\, \mathcal{L}(\bs{\theta}) \coloneqq \left(I - \bs{\theta}\bs{\theta}^\top\right) \nabla \mathcal{L} (\bs{\theta}),
$$
where $\nabla \mathcal{L} (\bs{\theta})$ is the Euclidean gradient of $\bs{\theta}$.

For example, the Riemannian gradient of $\bs{u}_{w}$ is computed as 
\begin{align*}
\text{grad}\, \mathcal{L}(\bs{u}_{w_j}) &=
\left(I - \bs{u}_{w_j}\bs{u}_{w_j}^\top\right) \bigg( \sum_{c\in \mathcal{T}}\mathds{1}(w_j\in \mathcal{C}) \kappa_{c}\bs{c} + \\
& \sum_{d_i, w_{j+k}} \mathds{1}(\text{pos}_{d_i, w_j, w_{j+k}}-\text{neg} < m) \left( \bs{v}_{w_{j+k}} + \bs{d}_i \right) \bigg), 
\end{align*}
where $\mathds{1}(w_j\in \mathcal{C})$ is the indicator function of whether $w_j$ belongs to category $c$; $\mathds{1}(\text{pos}_{d_i, w_j, w_{j+k}}-\text{neg} < m)$ is the indicator function of whether the margin of the positive tuple over the negative one is achieved.

The Riemannian gradient of the other embeddings can be derived similarly. Since we aim to maximize our objective, we update the parameters following the Riemannian gradient direction:
\begin{equation*}
\label{eq:rsg}
\bs{\theta}^{(t+1)} \gets R_{\bs{\theta}^{(t)}}\left(\alpha \cdot \text{grad}\, \mathcal{L}\left(\bs{\theta}^{(t)}\right)\right),
\end{equation*}
where $\alpha$ is the learning rate; $R_{\bs{x}}(\bs{z})$ is a first-order approximation of the exponential mapping at $\bs{x}$ which maps the updated parameters back to the sphere. We follow the definition in \cite{meng2019spherical}:
$$
R_{\bs{x}}\left(\bs{z} \right) \coloneqq \frac{\bs{x} + \bs{z}}{\|\bs{x} + \bs{z}\|}.
$$

\subsection{Overall Algorithm}
We summarize the overall algorithm of Hierarchical Topic Mining in Algorithm~\ref{alg:train}.
\begin{algorithm}[h]
\caption{Hierarchical Topic Mining.}
\label{alg:train}
\KwIn{
A text corpus $\mathcal{D}$; a category tree $\mathcal{T} = \{c_{i}\}|_{i=1}^{n}$; number of terms $K$ to retrieve per category .
}
\KwOut{Hierarchical Topic Mining results $\mathcal{C}_i|_{i=1}^{n}$.}

$\bs{u}_w, \bs{v}_w, \bs{d}, \bs{c} \gets $ random initialization on $\mathbb{S}^{p-1}$\;
$t \gets 1$\;
$\mathcal{C}_i^{(1)} \gets w_{c_i}|_{i=1}^{n}$ \Comment{initialize with category names}\;

\While{$t < K + 1$}  {
$t \gets t + 1$\;
// Representative term retrieval\;
$\mathcal{C}_i^{(t)}|_{i=1}^{n} \gets $ Eq.~\eqref{eq:e_step}\Comment{E-Step}\;
// Embedding training\;
$\bs{u}_w, \bs{v}_w, \bs{d}, \bs{c} \gets $ Eq.~\eqref{eq:emb}\Comment{M-Step}\;
}
\For{$i \gets 1$ to $n$} {
$\mathcal{C}_i^{(t)} \gets \mathcal{C}_i^{(t)} \setminus \{w_{c_i}\}$\Comment{exclude category names}\;
}
Return $\mathcal{C}_i^{(t)}|_{i=1}^{n}$\;
\end{algorithm}

\noindent
\textbf{Complexity.} We analyze the computation cost of our algorithm with respect to the tree size $n$. The tree embedding objective (Eq.~\eqref{eq:obj_tree_cp}) loops over every local tree $\mathcal{T}_r\in \mathcal{T}$ and every pair of sibling nodes in $\mathcal{T}_r$. Since the number of local trees is upper bounded by the number of total tree nodes, the complexity is $\mathcal{O}(nB^2)$ where $B$ is the maximum branching factor in $\mathcal{T}$. The text embedding objective (Eq.~\eqref{eq:obj_text}) pushes each representative term into the spherical sector centered around the category center vector, whose complexity is $\mathcal{O}(nK)$. Overall, our algorithm scales linearly with the tree size.

%% file: 5-sum.tex

%% file: 6-exp.tex

\section{Experiments}
In this section, we conduct empirical evaluations to demonstrate the effectiveness of our model. We also carry out case studies to show how the joint embedding space effectively models category tree structure and textual semantics.

\subsection{Experiment Setup}
\noindent
\textbf{Datasets.}
\begin{table}[t]
\centering
\caption{Dataset statistics.}
\label{tab:dataset_stats}
\scalebox{1.0}{
\begin{tabular}{cccc}
\toprule
Corpus & \# super-categories & \# sub-categories & \# documents\\
\midrule
\textbf{NYT} & 8 & 12 & 89,768 \\
\textbf{arXiv} & 3 & 29 & 230,105 \\
\bottomrule
\end{tabular}
}
\end{table}
We use two datasets from different domains with ground-truth category hierarchy: (1) The New York Times annotated corpus (\textbf{NYT}) \cite{Sandhaus2008}; (2) arXiv paper abstracts (\textbf{arXiv})\footnote{Data crawled from \url{https://arxiv.org/}.}.
For both datasets, we first select the major categories (with more than $1,000$ documents) and then collect documents with exactly one ground truth category label. The dataset statistics can be found at Table~\ref{tab:dataset_stats}.

\noindent
\textbf{Implementation Details and Parameters.}
We pre-process the corpora by discarding infrequent words that appear less than $5$ times. We use AutoPhrase~\cite{Shang2018AutomatedPM} to extract quality phrases, which are treated as single words during embedding training.
For fair comparisons with baselines, we set hyperparameters as below for all methods: Embedding dimension $p=100$; local context window size $h=5$; number of representative terms to retrieve per category $K=5$; learning rate $\alpha$ is set to be $0.025$ initially with linear decay. Other parameters (if any) are set to be the default values of the corresponding algorithm.

\subsection{Hierarchical Topic Mining}
\noindent
\textbf{Compared Methods.} We compare our model with the following baselines including unsupervised/seed-guided hierarchical topic models and unsupervised/seed-guided text embedding models. For baseline methods that require the number of topics $n_L$ at each level $L$ as input, we vary $n_L$ in $[n_L, 2n_L, \dots, 5n_L]$ where $n_L$ is the actual number of categories at level $L$ and report the best performance of the method.
\begin{itemize}[wide, labelwidth=!, labelindent=0pt] 
\item hLDA~\cite{Blei2003HierarchicalTM}: hLDA is a non-parametric hierarchical topic model. It assumes that documents are generated from the word distribution of a path of topics induced by the nested Chinese restaurant process. Since hLDA is unsupervised and cannot take given category names as supervision, we manually match the most relevant topics to the provided category hierarchy.
\item hPAM~\cite{Mimno2007MixturesOH}: hPAM generalizes the Pachinko Allocation Model~\cite{li2006pachinko} by sampling topic paths from the Dirichlet-multinomial distributions of internal nodes. We perform manual matching of topics as we do for hLDA.
\item JoSE~\cite{meng2019spherical}: JoSE trains spherical text embeddings with Riemannian optimization. It outperforms Euclidean embedding models on textual similarity measurement. We retrieve the nearest-neighbor words of the category name in the spherical space as category representative words.
\item Poincar{\'e} GloVe~\cite{Tifrea2019PoincareGH}: Poincar{\'e} GloVe learns hyperbolic word embeddings based on the Euclidean GloVe model. It naturally encodes the latent hierarchical word semantic correlations (\eg, hypernym-hyponym). We retrieve the nearest-neighbor words of the category name in the Poincar{\'e} space as category representative words.
\item Anchored CorEx~\cite{Gallagher2017AnchoredCE}:
CorEx discovers informative topics via total correlation maximization and can naturally model topic hierarchy via latent factor dependencies. Its anchored version incorporates user-provided seed words by balancing between compressing the original corpus and preserving anchor words related information. We provide the category names as seed words.
\item CatE~\cite{Meng2020DiscriminativeTM}: CatE takes category names as input and learns discriminative text embeddings by enforcing distinctiveness among categories. We recursively run CatE on local trees since CatE assumes that the provided categories are mutually-exclusive semantically.
\end{itemize}

\noindent
\textbf{Quantitative Evaluation.}
We apply two metrics on the top-$K$ ($K=5$ in our experiments) words/phrases retrieved under each category to evaluate all methods: Topic coherence (TC) and Mean accuracy (MACC) as defined in~\cite{Meng2020DiscriminativeTM}. The accuracy metric is obtained from the averaged results given by five graduate students who independently label whether each retrieved term is highly relevant to the corresponding category. The quantitative results are reported in Table~\ref{tab:main_results}.

\begin{table}[t]
\centering
\caption{Quantitative evaluation: Hierarchical Topic Mining.}
\label{tab:main_results}
\scalebox{1}{
\begin{tabular}{c|cc|cc}
\toprule
\multirow{2}{*}{Models} &
\multicolumn{2}{c|}{\textbf{NYT}} & \multicolumn{2}{c}{\textbf{arXiv}} \\
& TC & MACC & TC & MACC \\
\midrule
hLDA               & -0.0070         & 0.1636 & -0.0124 & 0.1471 \\
hPAM               &  0.0074         & 0.3091 &  0.0037 & 0.1824 \\
JoSE               &  0.0140         & 0.6818 &  0.0051 & 0.7412\\
Poincar{\'e} GloVe &  0.0092         & 0.6182 & -0.0050 & 0.5588 \\
Anchored CorEx     &  0.0117         & 0.3909 &  0.0060 & 0.4941 \\
CatE               &  0.0149         & 0.9000 &  0.0066 & 0.8176 \\
\josh               & \textbf{0.0166} & \textbf{0.9091} &  \textbf{0.0074} & \textbf{0.8324} \\
\bottomrule
\end{tabular}
}
\end{table}

\noindent
\textbf{Qualitative Results.}
\begin{table}[t]
\centering
\caption{Run time (in minutes) on \textbf{NYT}. Models are run on a machine with $20$ cores of Intel(R) Xeon(R) CPU E5-2680 v2 @ $2.80$ GHz.}
\label{tab:time}
\scalebox{0.85}{
\begin{tabular}{*{7}{c}}
\toprule
hLDA & hPAM & JoSE & Poincar{\'e} GloVe & Anchored CorEx & CatE & \josh \\
\midrule
53 & 22 & 5 & 16 & 61 & 52 & 6 \\
\bottomrule
\end{tabular}
}
\vspace*{-1em}
\end{table}
We demonstrate the qualitative results of \textbf{NYT} in Figure~\ref{fig:nyt} and \textbf{arXiv} in Figure~\ref{fig:arxiv} in Appendix~\ref{sec:arxiv}. Words in blue boxes are input category names; words in white boxes are retrieved representative terms of the corresponding category.

\begin{figure*}[t]
\centering
\includegraphics[width=\linewidth]{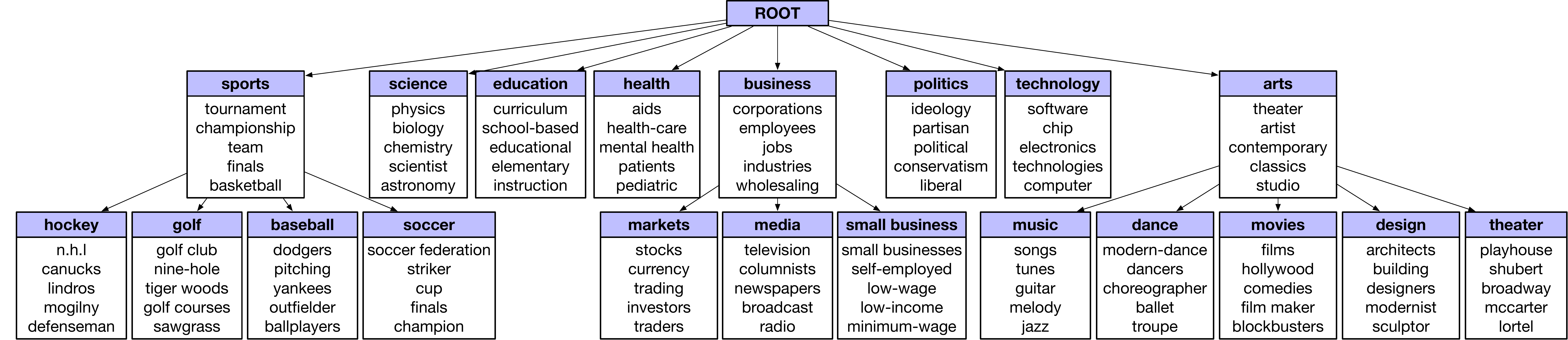}
\caption{Hierarchical Topic Mining results on \textbf{NYT}.}
\label{fig:nyt}
\end{figure*}

\noindent
\textbf{Run Time.}
Since topic discovery is usually performed on large-scale text corpus, algorithm efficiency is of great importance. Therefore, we report the run time of all methods in Table~\ref{tab:time}. \josh takes only slightly longer to train than JoSE, which only learns text embeddings.

\noindent
\textbf{Discussions.} The two unsupervised baselines (hLDA and hPAM) do not perform well. Despite running with different parameters for multiple times, they still fail to generate high quality topics similar to the ground-truth category hierarchy, showing the limitations of unsupervised approaches. For the two unsupervised embedding baselines, JoSE outperforms Poincar{\'e} GloVe by a large margin, demonstrating that the spherical space is more suitable than the hyperbolic space on capturing textual semantic correlations for category representative term retrieval. CatE has strong performance on the two datasets, but it has to be run recursively on each set of sibling nodes since it requires all the input categories to be mutually exclusive. Therefore, run time will become a potential bottleneck of applying CatE to large-scale hierarchies. \josh not only outperforms all models on Hierarchical Topic Mining quality, but also enjoys high efficiency via efficient joint modeling of category tree structure and text corpus statistics.

\subsection{Weakly-Supervised Hierarchical Text Classification}
Hierarchical Topic Mining is also closely related to the task of text classification. Intuitively, having a good understanding of topics should lead to better categorization of documents. Similar to Hierarchical Topic Mining, the input to weakly-supervised hierarchical classification is also a word-described category tree. Since weakly-supervised classification~\cite{Meng2018WeaklySupervisedNT,Meng2019WeaklySupervisedHT,zhang2019higitclass} does not require training documents, it is especially favorable when manual annotation is expensive. 

\noindent
\textbf{Compared Methods.} We compare the following weakly-supervised hierarchical models on their classification performance, evaluated on the two datasets.

\begin{itemize}[wide, labelwidth=!, labelindent=0pt]
\item
WeSHClass~\cite{Meng2019WeaklySupervisedHT}: WeSHClass leverages the provided keywords of each category to generate a set of pseudo documents for pre-training a hierarchical deep classifier, and self-trains the ensembled local classifiers  on unlabeled data. It uses Word2Vec~\cite{Mikolov2013DistributedRO} as word representation.
\item
\josh: Since our model makes explicit generative assumption between topics and documents (Eq.~\eqref{eq:gen_topic}), we are able to build a generative classifier by assigning the document to the category with the highest probability that it gets generated from, \ie,
$$
y_d = \argmax_c \text{vMF}(\bs{d};\bs{c},\kappa_{c}),
$$
where $y_d$ is the predicted category label for document $d$.
\item 
WeSHClass + CatE~\cite{Meng2020DiscriminativeTM}: It is shown in \cite{Meng2020DiscriminativeTM} that the learned discriminative text embedding can be used as input feature to benefit classification model. We replace the Word2Vec embedding used in WeSHClass with CatE embeddings.
\item 
WeSHClass + \josh: We replace the Word2Vec embedding used in WeSHClass with word embeddings learned by \josh. Since \josh effectively leverages the category tree structure to guide text embedding configuration, it is expected to benefit hierarchical classification model as input features.
\end{itemize}

\noindent
\textbf{Quantitative Evaluation.}
We use two metrics for classification evaluation, Macro-F1 and Micro-F1, which are commonly used in multi-class classification evaluations. The results are reported in Table~\ref{tab:class_results}.

\noindent
\textbf{Discussions.} We demonstrate two potential usage of \josh in weakly-supervised hierarchical text classification: (1) Directly build a generative classifier based on the model assumption; (2) Use the learned embedding as input features to existing classification models. \josh alone as a generative classifier even outperforms the WeSHClass model;  when used as features to WeSHClass, \josh significantly boosts the classification performance,
proved to be more effective than CatE which does not model the category hierarchy.

\begin{table}[t]
\centering
\caption{Quantitative evaluation: Weakly-supervised hierarchical classification.}
\vspace*{-1em}
\label{tab:class_results}
\scalebox{0.93}{
\begin{tabular}{c|cc|cc}
\toprule
\multirow{2}{*}{Models} &
\multicolumn{2}{c|}{\textbf{NYT}} & \multicolumn{2}{c}{\textbf{arXiv}} \\
& Macro-F1 & Micro-F1 & Macro-F1 & Micro-F1 \\
\midrule
WeSHClass          & 0.425 & 0.581  &  0.320 &  0.542  \\
\josh               &  0.429  & 0.600 &  0.367 & 0.610 \\
\midrule
WeSHClass + CatE  & 0.503 & 0.679 & 0.401 & 0.622 \\
WeSHClass + \josh & \textbf{0.582} & \textbf{0.703} & \textbf{0.412} & \textbf{0.673} \\
\bottomrule
\end{tabular}
}
\vspace*{-1em}
\end{table}

\subsection{Joint Embedding Space Visualization}
\begin{figure*}[t]
\centering
\subfigure[\textbf{NYT} joint embedding space.]{
\label{fig:nyt_vis}
\includegraphics[width = 0.485\linewidth]{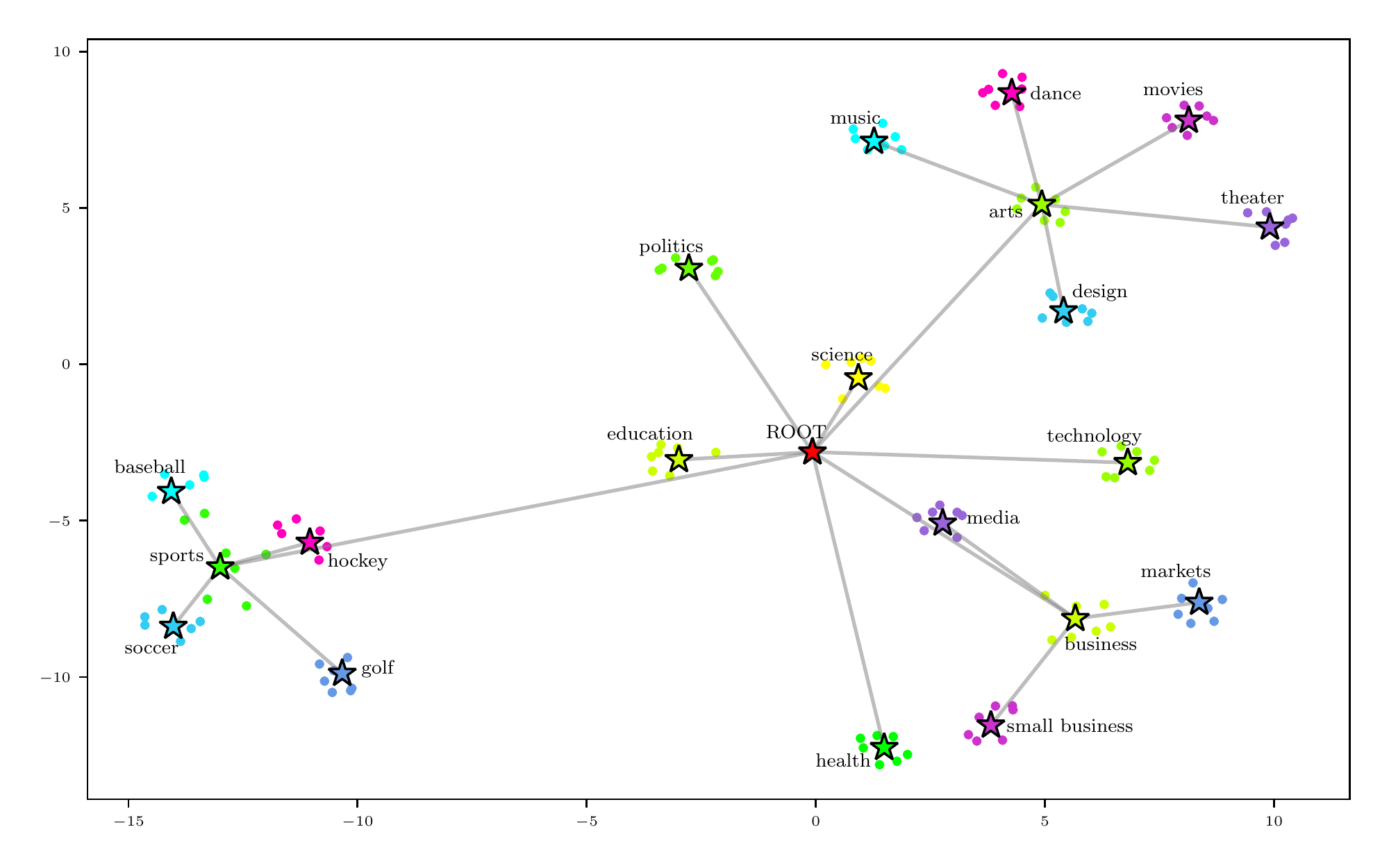}
}
\subfigure[\textbf{arXiv} joint embedding space.]{
\label{fig:arxiv_vis}
\includegraphics[width = 0.485\linewidth]{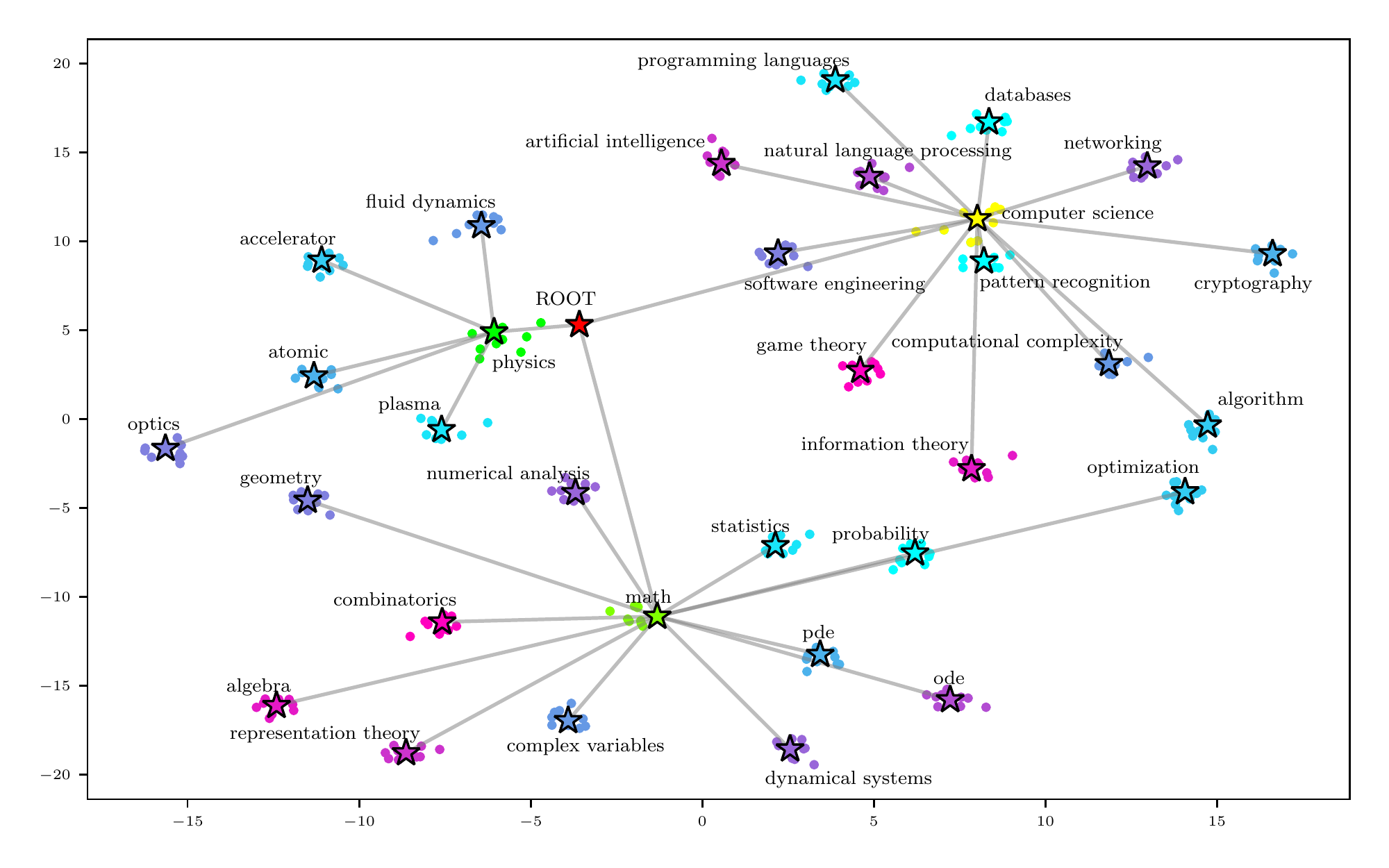}
}
\vspace{-1em}
\caption{Joint embedding space visualization. Category center vectors are denoted as stars; representative words are denoted as dots in the same color with corresponding category.}
\label{fig:vis}
\vspace{-1em}
\end{figure*}
To understand how categories and words are distributed in the joint embedding space and how the category tree structure is modeled, we apply t-SNE~\cite{Maaten2008VisualizingDU} to visualize the embedding space in Figure~\ref{fig:vis}. Representative terms surround their category centers; sub-categories surround their super-categories which form a category tree structure. An interesting observation is that some sub-categories under different super-categories are embedded closer, \eg, in Figure~\ref{fig:arxiv_vis}, ``optimization'' under ``math'' and ``algorithm'' under ``computer science''. Indeed, these two sub-categories are somewhat cross-domain---``optimization'' and ``algorithm'' are relevant to both mathematics and computer science. This shows that \josh
not only models the given category tree structure, but also captures semantic correlation among categories via jointly training tree and text embedding.

%% file: 7-related.tex

\section{Related Work}

\subsection{Hierarchical Topic Modeling}
Hierarchical topic models extend their flat counterparts by capturing the correlations among topics and generate topic hierarchies. hLDA~\cite{Blei2003HierarchicalTM} generalizes LDA~\cite{Blei2003LatentDA} with a non-parametric probabilistic model, the nested Chinese restaurant
process, which induces a path from the root topic to a leaf topic. The documents are assumed to be generated by sampling words from the topics along this path. Another famous hierarchical topic model, hPAM ~\cite{Mimno2007MixturesOH} is built on  the Pachinko Allocation Model~\cite{li2006pachinko} which models documents as a mixture of distributions over a set of topics; the co-occurrences of topics are further represented via a directed acyclic graph. hPAM represents the topic hierarchical structure through the Dirichlet-multinomial parameters of the internal node distributions. There are also supervised hierarchical topic models. 
HSLDA~\cite{Perotte2011HierarchicallySL} extends sLDA~\cite{mcauliffe2008supervised} by incorporating a breadth first traversal in the label space during document generation. 
SSHLDA~\cite{Mao2012SSHLDAAS} is a semi-supervised hierarchical topic model that not only explores new latent topics in the label space, but also makes
use of the information from the hierarchy of observed labels. 
A seed-guided topic modeling framework, CorEx~\cite{Gallagher2017AnchoredCE}, learns informative topics that maximize total correlation. It is similar to our setting as it incorporates seed words by preserving seed relevant information. CorEx is able to generate topic hierarchy via latent factor dependencies. Different from the previous unsupervised and supervised topic models, our framework takes as guidance only a category hierarchy described by category names, and models category-word semantic correlation via joint spherical text and tree embedding.

\subsection{Text Embedding and Tree Embedding}
Text embeddings~\cite{Le2014DistributedRO,Mikolov2013DistributedRO,meng2019spherical,meng2020unsupervised,Pennington2014GloveGV} effectively capture textual semantic similarity via distributed representation learning of words, phrases, sentences, etc. Several topic modeling frameworks, such as~\cite{batmanghelich2016nonparametric,Dieng2019TopicMI,Liu2015TopicalWE} leverage text embeddings to model contextualized semantic similarity of words, making up the bag-of-words generative assumption in classical topic models.
Poincar{\'e} GloVe~\cite{Tifrea2019PoincareGH} adapts the original GloVe model by training word embedding in the Poincar{\'e} space where the latent hierarchical semantic relations between words are naturally captured.
A recent text embedding model CatE~\cite{Meng2020DiscriminativeTM} proposes to learn discriminative text embeddings for category representative term retrieval given a set of category names as user guidance, which is similar to our setting. CatE makes mutual exclusive assumption on category semantics, which does not hold when categories exhibit a hierarchical structure. None of the previous text embedding framework is able to model a given hierarchical category structure in the embedding space to guide text embedding learning.

With the recent advances in hyperbolic embedding space, several frameworks have been developed to model tree structures. Poincar{\'e}  embedding~\cite{Nickel2017PoincarEF} learns to model hierarchical structure in the Poincar{\'e} ball. Since the embedding distance directly corresponds to tree distance, Poincar{\'e} embedding can be used to infer lexical entailment relationship by embedding the tree structure of WordNet or perform link prediction by embedding networks.
Later, Lorentz model~\cite{Nickel2018LearningCH} brings a more principled optimization approach in the hyperbolic space to learn tree structures; hyperbolic cones~\cite{Ganea2018HyperbolicEC} are proposed to model hierarchical relations
and admit an optimal shape with a closed form expression. 
These hyperbolic tree embedding methods, however, are not suitable for embedding category trees in a joint space with words. The reason is that hyperbolic embeddings preserve the \emph{absolute} tree distance, \ie, similar embedding distances imply similar tree distances. In a category tree, lower-level sibling categories are generally more semantically similar than higher-level ones despite the same tree distance. Therefore, category embedding distances should not be solely determined by tree distances. In our model, text and category tree are jointly embedded, allowing the tree structure to better reflect the textual semantics of the categories.

%% file: 8-concl.tex

\section{Conclusions and Future Work}
In this paper, we propose a new task for hierarchical topic discovery guided by a user-provided category tree described with category names only. To effectively model the category tree structure while capturing text corpus statistics, we propose a joint spherical space embedding model \josh that uses directional similarity to characterize semantic correlations among words, documents, and categories. We develop an EM algorithm based on Riemannian optimization for training the model in the spherical space. \josh mines high-quality topics and enjoys high efficiency. We also show that \josh can be applied to the task of weakly-supervised hierarchical classification, serving as either a generative classifier on its own, or input features to existing classification models.

In the future, we aim to extend \josh to not only focus on a user-given category structure, but also be able to discover other latent topics from a text corpus, probably by relaxing the assumption that a document is generated from one of the given topics or collaborating with other taxonomy construction algorithms~\cite{Huang2020CoRel,Huang2020GuidingCS}. Also, the promising results of our joint spherical space embedding model may shed light on future studies of embedding tree or graph structures along with textual data in the spherical space for mining structured knowledge from text corpora.

%% file: 9-app.tex

\clearpage
\appendix

\section{Hierarchical Topic Mining Results on arXiv}
\label{sec:arxiv}

\begin{figure}[h]
	\centering
	\subfigure[``Math'' subtree.]{
		\label{fig:arxiv_math}
		\includegraphics[width = \linewidth]{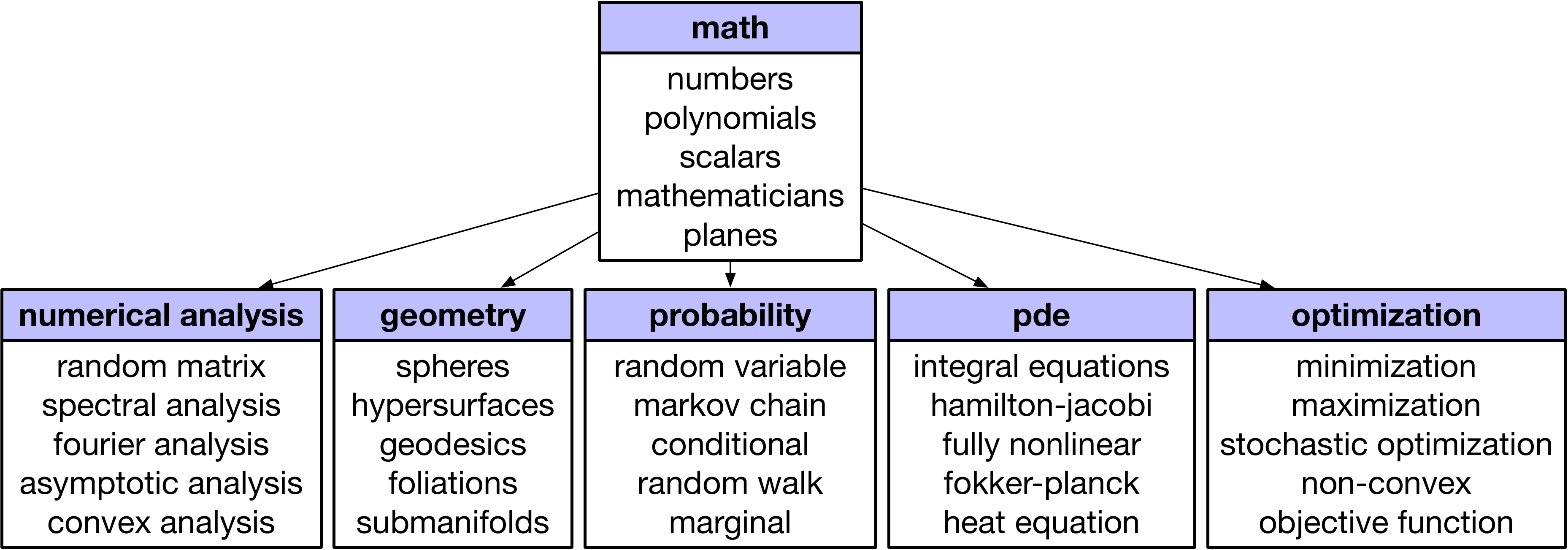}
	}
	\subfigure[``Physics'' subtree.]{
		\label{fig:arxiv_physics}
		\includegraphics[width = \linewidth]{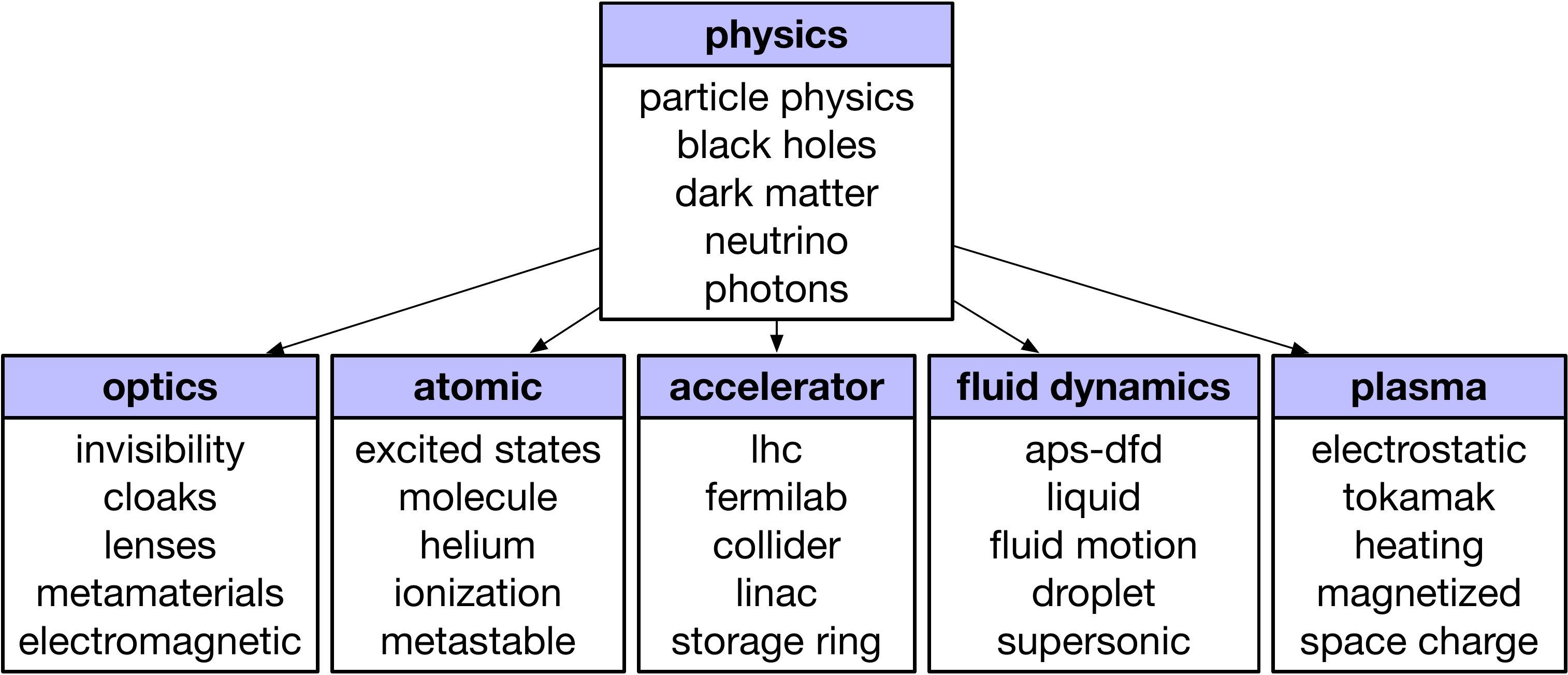}
	}
	\subfigure[``Computer Science'' subtree.]{
		\label{fig:arxiv_cs}
		\includegraphics[width = \linewidth]{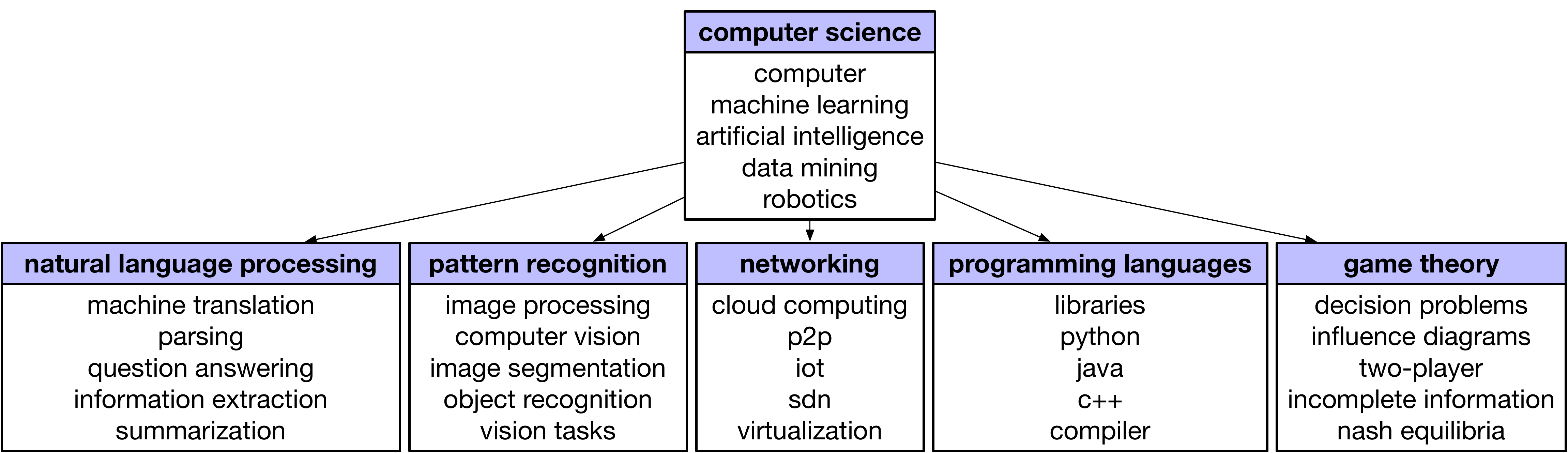}
	}
	\vspace*{-2ex}
	\caption{Results of Hierarchical Topic Mining on \textbf{arXiv}: Only 5 sub-categories per super-category are shown here.}
	\label{fig:arxiv}
\end{figure}

Figure~\ref{fig:arxiv} shows part of the Hierarchical Topic Mining results on \textbf{arXiv}. 
\vspace*{-3.2ex}

\vspace*{-2ex}
\section{Derivation of Objective}
\label{sec:app_obj}


The derivation of Eqs.~\eqref{eq:obj_tree_cp} and \eqref{eq:obj_text} is provided as follows.

The conditional likelihood of the corpus given the category hierarchy is obtained by combining the assumptions described in Eqs.~\eqref{eq:gen_topic_mod}, \eqref{eq:gen_global} and \eqref{eq:gen_local}:
\begin{equation}
\begin{aligned}
\label{eq:likelihood}
P(\mathcal{D} \mid \mathcal{T}) &= \prod_{d_i\in \mathcal{D}} p(d_i \mid c_i) \prod_{w_j \in d_i} p(w_j \mid d_i) \prod_{\substack{w_{j+k} \in d \\ -h \le k \le h, k \neq 0}} p(w_{j+k} \mid w_j) \\
&\propto \prod_{d_i\in \mathcal{D}} \prod_{w_j \in d_i} p(w_j\mid c_i) p(w_j \mid d_i) \prod_{\substack{w_{j+k} \in d_i \\ -h \le k \le h, k \neq 0}} p(w_{j+k} \mid w_j),
\end{aligned} 
\end{equation}
where $c_i$ is the latent true category of $d_i$.

To make the learning of text embedding and category distribution explicit, we re-write Eq.~\eqref{eq:likelihood} by re-arranging the product of $p(w\mid c)$ over categories:
\begin{equation*}
\begin{aligned}
\label{eq:re-likelihood}
P(\mathcal{D} \mid \mathcal{T})
&\propto \prod_{c_i \in \mathcal{T}} \prod_{w_j \in c_i} p(w_j\mid c_i)  \\
& \quad \cdot \prod_{d_i\in \mathcal{D}} \prod_{w_j \in d_i}  p(w_j \mid d_i) \prod_{\substack{w_{j+k} \in d_i \\ -h \le k \le h, k \neq 0}} p(w_{j+k} \mid w_j),
\end{aligned} 
\end{equation*}

Taking the log-likelihood as our objective to maximize, we have
\begin{equation}
\begin{aligned}
\label{eq:log-like}
\mathcal{L} &= \sum_{c_i \in \mathcal{T}} \sum_{w_j \in c_i} \log p(w_j\mid c_i)  \\
&\quad +\sum_{d_i\in \mathcal{D}} \sum_{w_j \in d_i} \log p(w_j \mid d_i)\\
&\quad + \sum_{d_i\in \mathcal{D}} \sum_{w_j \in d_i} \sum_{\substack{w_{j+k} \in d_i \\ -h \le k \le h, k \neq 0}} \log p(w_{j+k} \mid w_j)\\
&\quad + \text{constant}.
\end{aligned} 
\end{equation}

We omit the constant term and split Eq.~\eqref{eq:log-like} into category distribution modeling and corpus-based embedding learning objectives, plugging in the definition of the probability expressions given by Eqs.~\eqref{eq:gen_topic_mod}, \eqref{eq:gen_global} and \eqref{eq:gen_local}. For category distribution modeling, we have:
\begin{equation}
\begin{aligned}
\label{eq:cat}
\mathcal{L}_{\text{cat}} &= \sum_{c_i \in \mathcal{T}} \sum_{w_j \in c_i} \log p(w_j\mid c_i) \\
&= \sum_{c_i \in \mathcal{T}} \sum_{w_j \in c_i} \log \left(n_p(\kappa_{c_i})\right) + \kappa_{c_i} \cdot \cos(\bs{u}_{w_j}, \bs{c}_i).
\end{aligned} 
\end{equation}
Eq.~\eqref{eq:cat} achieves the same effect as Eq.~\eqref{eq:obj_intra} on encouraging word representative terms to have high directional similarity with the category center vector, except that Eq.~\eqref{eq:cat} does not incorporate an intra-category margin. Thus we extend Eq.~\eqref{eq:cat} into the following:
\begin{equation}
\begin{aligned}
\label{eq:obj_cat_mod}
\mathcal{L}_{\text{cat}}^* &= \sum_{c_i \in \mathcal{T}} \sum_{w_j \in \mathcal{C}_i} \left( \log \left(n_p(\kappa_{c_i})\right) + \kappa_{c_i} \bs{u}_{w_j}^\top \bs{c}_i \right) \mathds{1}(\bs{u}_{w_j}^\top \bs{c}_i < m_{\text{intra}}),
\end{aligned} 
\end{equation}
where $\mathds{1}(\cdot)$ is the indicator function.

For corpus-based embedding learning, we have:
\begin{align*}
\mathcal{L}_{\text{corpus}} &= \sum_{d_i\in \mathcal{D}} \sum_{w_j \in d_i} \left(\log p(w_j \mid d_i) + \sum_{\substack{w_{j+k} \in d_i \\ -h \le k \le h, k \neq 0}} \log p(w_{j+k} \mid w_j)\right)\\
&= \sum_{d_i\in \mathcal{D}} \sum_{w_j \in d_i} \left( \cos(\bs{u}_{w_j}, \bs{d}_i) + \sum_{\substack{w_{j+k} \in d_i \\ -h \le k \le h, k \neq 0}} \cos(\bs{v}_{w_{j+k}}, \bs{u}_{w_j}) \right).
\end{align*}

Directly maximizing the above objective results in trivial solution that all text embedding vectors are converged to the same point (so that the cosine similarity term is always maximized). To tackle this issue, we employ the same technique used in \cite{meng2019spherical} where the log-likelihood of a positive co-occurring tuple $(w_j, w_{j+k}, d_i)$ is pushed over that of a negative tuple $(w_j', w_{j+k}, d_i)$ by a margin $m$, where $w_j'$ is a randomly sampled word from the vocabulary. 
\begin{align*}
\label{eq:obj_text_mod}
\mathcal{L}_{\text{corpus}} &= \sum_{d_i\in \mathcal{D}} \sum_{w_j \in d_i} \sum_{\substack{w_{j+k} \in d_i \\ -h \le k \le h, k \neq 0}} \min \bigg(0, -m + \cos(\bs{v}_{w_{j+k}}, \bs{u}_{w_j}) + \\
& \qquad \cos(\bs{u}_{w_j}, \bs{d}_i)) - \cos(\bs{v}_{w_{j+k}}, \bs{u}_{w_j'}) - \cos(\bs{u}_{w_j'}, \bs{d}_i) \bigg).
\end{align*}
Finally, $
\mathcal{L}_{\text{text}} = \mathcal{L}_{\text{cat}}^* + \mathcal{L}_{\text{corpus}}.
$